%% file: main.tex
\documentclass{article} 
\usepackage{colm2024_conference}

\usepackage{microtype}
\usepackage{hyperref}
\usepackage{url}
\usepackage{booktabs}

\usepackage{graphicx}
\usepackage{CJKutf8}
\usepackage{multirow}

\colmfinalcopy

\title{Code-Based English Models Surprising Performance on Chinese QA Pair Extraction Task}

\author{Linghan Zheng\footnotemark[2], Hui Liu\footnotemark[2]\thanks{Corresponding author: hui.liuhui@antgroup.com},Xiaojun Lin\thanks{Hui Liu completed the early experiments and discovered the outperformance of code-based models. Linghan Zheng conducted additional experiments to supplement the findings, and Xiaojun Lin distilled the paper's perspectives and crystallized the conceptual insights. Equal contribution.}, Jiayuan Dong, Yue Sheng, Gang Shi, Zhiwei Liu, Hongwei Chen
\\
Ant Group, China
}

\begin{document}

\maketitle

\begin{abstract}
In previous studies, code-based models have consistently outperformed text-based models in reasoning-intensive scenarios. When generating our knowledge base for Retrieval-Augmented Generation (RAG), we observed that code-based models also perform exceptionally well in Chinese QA Pair Extraction task. Further, our experiments and the metrics we designed discovered that code-based models containing a certain amount of Chinese data achieve even better performance. Additionally, the capabilities of code-based English models in specified Chinese tasks offer a distinct perspective for discussion on the philosophical "Chinese Room" thought experiment.

\end{abstract}

\section{Introduction}
\input{introduction}

\section{Related Work}
\input{related_work}

\section{Methodology}
\input{methodology}
\label{sec:methodology}

\section{Experiments}
\input{evaluation}

\section{Discussion}
While numerous studies indicate that the incorporation of code data significantly enhances the reasoning abilities of LLMs~\citep{Liang2023HolisticEO, Wang2022Code4StructCG}, there is a contention among some scholars that such improvements are more evident in smaller-scale LLMs. They suggest that for exceedingly large models, the complex reasoning capabilities occur through the process of ability sharing across a diverse range of data sources, diminishing the impact of code data on augmenting reasoning faculties. We are particularly interested in understanding the performance deltas that may exist between code-augmented models and their counterparts when model sizes increase. 

A growing body of work suggests that LLMs do not require all fitted parameters to retain their learned hypotheses. In this paper, we speculate that with an English-centric, code-based model performing Chinese tasks, it may be possible to remove over 99 percent of the model's weights without any significant degradation in performance.

\section{Conclusions}

In this paper, we categorize various LLM tasks into two types: one that requires the LLM to provide additional knowledge, such as translation, and another that entails data processing tasks with given materials, such as summarization, knowledge extraction, and comprehension based on provided texts. Within the second category, we describe a Chinese knowledge generation task that involves generating question-answer pairs from material. Through experiments with multiple models and human evaluation, we found that a code-based model with limited understanding of Chinese outperformed the other models. To explain this outcome, we constructed several statistical metrics, including "text extraction aggregation count" and "content creation rate". We discovered that the code-based model's strong ability to "Copy The Original Text" is the reason for its superior performance in Chinese tasks. We believe this finding could be applicable to many other Chinese language tasks. 

We propose that our approach to evaluating models in this task completely eliminates the risk of test set "contamination" from the training data. By impairing the model's Chinese capabilities through vocabulary changes and subsequently training it with a small, identical amount of Chinese data for evaluation, we can effectively assess the performance differences between models. This method allows for the reuse of existing Chinese benchmarks~\citep{Li2023CLEVACL} to build a variety of evaluation tools, providing a fresh approach to the challenging problem of how to evaluate large models.

\section{Future Work}

\subsection{Proposing a Controlled AI Framework Using a Regulated Chinese Manual}
\begin{figure}[ht]
    \begin{center}
    \includegraphics[width=0.5\linewidth]{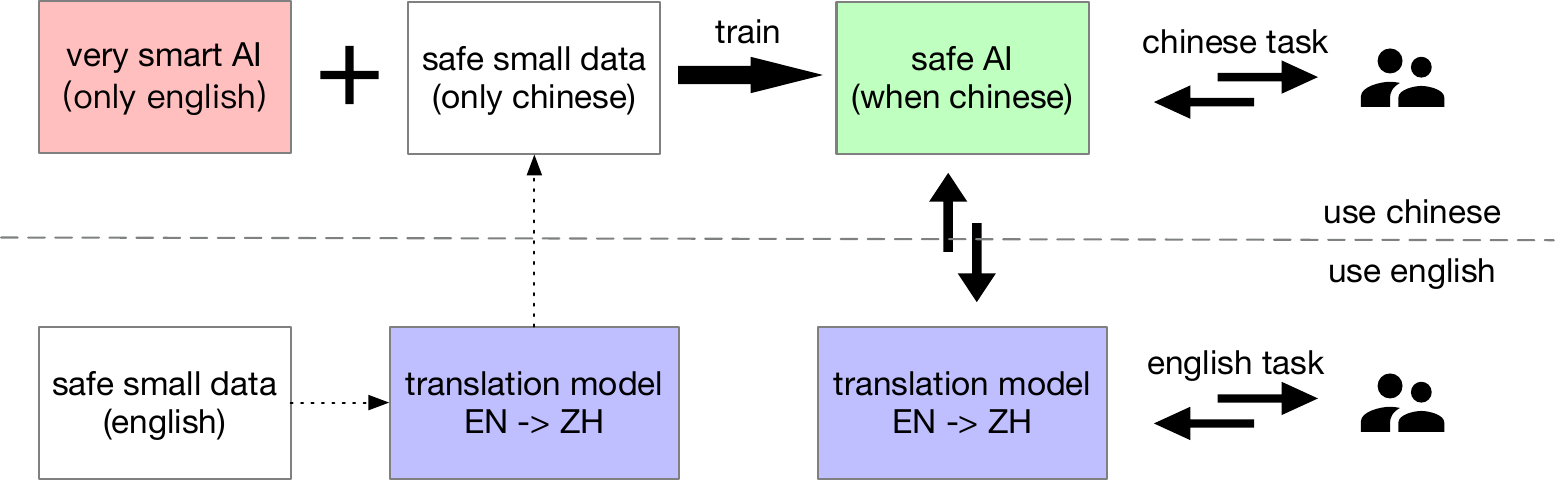}
    \caption{New Chinese Room}
    \label{fig:new_chinese_room}
    \end{center}
\end{figure}

In future work, we aim to construct a controlled AI framework inspired by the "Chinese Room" thought experiment~\citep{Moural2003JohnST}. Utilizing an English code model that has limited understanding of Chinese, we plan to explore the development of a system that performs Chinese language tasks with exceptional skill without relying on translation or dictionaries. This AI framework will operate within the confines of a regulated Chinese manual, serving as a safeguard for task-specific applications and prompting further philosophical and practical inquiries into AI constraints. As shown in figure \ref{fig:new_chinese_room}.

Given the advancement towards potential superhuman AI models~\citep{Burns2023WeaktoStrongGE}, we advocate for a controlled AI framework. We suggest that a tightly regulated Chinese manual can effectively constrain the AI to specific tasks, acting as a control mechanism. This proposition urges philosophers and AI researchers to further investigate the practicality of such a system.

\subsection{Future Research Directions Involving Code-based Models for Chinese-related Tasks}
Building on the initial findings, our future research will investigate the broader potential of code-based models in handling Chinese-related tasks. We will tackle challenges such as the interpretation of garbled Chinese characters and strive to enhance OCR capabilities for Chinese texts. Our goal is to utilize the distinctive strengths of code-based models to improve performance across various tasks within the Chinese language domain, potentially transforming the landscape of language-processing AI.

\bibliography{colm2024_conference}
\bibliographystyle{colm2024_conference}

\appendix
\section{Appendix}
\input{appendix}

\end{document}

%% file: introduction.tex
Recent advancements in the field of natural language processing (NLP) have led to the development of increasingly sophisticated models capable of understanding and generating human language with significant proficiency. Scaling up the size of language models has been shown to confer a range of benefits, such as improved performance and sample efficiency\citep{Kaplan2020ScalingLF}. Fine-tuning large models for diverse scenarios has also become consensus practice in the community.Traditionally, language models and code-based models\citep{code_llama,Feng2020CodeBERTAP} have been separated into distinct categories based on their domains of expertise, with the former excelling in general linguistic tasks and the latter in programming-related scenarios. However, an interesting observation has arisen in our experiments with Chinese text data generation tasks—intuitively, one would expect such tasks to be dominated by Chinese domain-based language models, but code-based models trained on English datasets have, in fact, exhibited superior performance.

This unexpected discovery challenges the traditional view that pre-trained models are domain-specific and calls for a more in-depth examination of their capabilities beyond their primary training language or format. The effectiveness of code-based models in Chinese text data generation tasks implies that the foundational skills learned during training are not strictly language-bound but may be transferable across different linguistic contexts. Such an insight necessitates a reevaluation of the current transfer learning paradigm that emphasizes language consistency in the selection of pre-trained models. It suggests that task-relevant skills and structural understanding might be more important than linguistic congruence in determining a model's applicability. This aligns with a broader theme in machine learning, where models are abstractions over tasks, and it is the nature of the task, rather than the form of the data, that dictates the most effective learning approach.

This paper focuses on dissecting the unexpected effectiveness of code-based models over language models in the context of Chinese text data generation tasks and to explore the broader implications of this finding for the NLP field. We hypothesize that the proficiency of code-based models in this domain may stem from their inherent ability to handle structured data, which aligns with the requirements of data generation tasks. In Section 3, we provide a detailed definition of the specific tasks, along with some nuances of the experimental and training processes. In Section 4, we present all experimental results and discuss the potential reasons behind the counterintuitive outcomes. This work offers insights into the essence of pre-trained models and their application in tasks that may not seem immediately related but actually share common underlying structures.

%% file: related_work.tex
In the wake of our recent findings, we identify several research directions that are intertwined with our discoveries, of which two stand out as perhaps the most relevant. Here, we describe these two directions and some pertaining literature that are closely associated with our observations.

The first pertinent research direction involves the transfer of monolingual models to new languages. This line of inquiry explores the adaptation of models trained solely in one language, such as English, to another language, like Chinese, while maintaining or enhancing performance. This transfer typically requires overcoming structural and lexical disparities between languages. ~\citet{Lample2019CrosslingualLM} introduces the Cross-lingual Language Model (XLM), which is a multilingual pretraining approach that facilitates cross-lingual transfer by learning shared representations across multiple languages. ~\citet{Artetxe2019OnTC} proposes an alternative approach that transfers a monolingual model to new languages at the lexical level and achieves competitive performance with multilingual BERT on standard cross-lingual classification benchmarks and a newly introduced Cross-lingual Question Answering Dataset (XQuAD).

The second research direction concerns cross-domain learning. This approach looks at how to transfer models from their original training domain, such as text processing, to a new domain, like code generation or converting text to code. Such transfers necessitate the models to understand the relationships between different domains and to be able to interchange information across them.  ~\citet{Thawani2021RepresentingNI} surveyed numerical understanding in NLP models using 7 sub-tasks such as measurement estimation and word problems.~\citet{Vulic2020ProbingPL} conducted on six typologically diverse languages and five distinct lexical tasks reveals consistent patterns and best practices that are universally applicable. However, the study also identifies notable variations across languages and tasks.
~\citet{Feng2020CodeBERTAP} present CodeBERT, a model pre-trained on both natural language and programming language that shows impressive capabilities in code understanding and generation tasks. 

Both of these research directions resonate with our observed phenomenon that code-based models may outperform language models in tasks that are not in the models' original training language.

%% file: methodology.tex
To prepare data for Retrieval-Augmented Generation (RAG)~\citep{Lewis2020RetrievalAugmentedGF}, we extract multiple question-answer pairs containing knowledge points from the given material. These pairs serve as a training dataset for the RAG model, enabling models to learn how to generate questions that are contextually relevant and to retrieve the corresponding answers from the document. The extraction process involves using algorithms capable of understanding the text's structure and content, identifying key facts, concepts, and themes that can form the basis of meaningful questions. During the extraction process, it is essential to ensure that the pairs meet the requirements of being organized, precise, informative, and exhibit lower hallucination, and that there is minimal overlap in the content covered by each pair to maintain diversity.

In this section, we describe the experiments we conducted to automatically generate high-quality question-answer pairs from our private documents. We first introduce the datasets we use in Section \ref{subsec3.1}. We then explain the models and the fine-tuning process used in our method with implementation details in Section \ref{subsec3.2}.

\subsection{Datasets}\label{subsec3.1}

Generating question-answer pairs from original documents has many uses, such as being utilized in RAG. Question answering (QA) is a central and highly popular area in NLP. There are currently an abundance of datasets available about QA generating, including ~\cite{Lewis2019MLQAEC} and ~\cite{Shao2018DRCDAC}. The first dataset comprised multilingual data; however, the data quality was relatively low. Given our research focus on the Chinese language domain, we decided against utilizing this dataset. The second dataset we evaluated contained predominantly Chinese-language data. Despite this, the structural complexity of the problems was limited. The dataset largely consisted of direct replications of existing document knowledge, primarily simple knowledge points such as numbers and dates, with scant evidence of complex knowledge extraction and generative outcomes. In light of these limitations, we resolved to create our own high-caliber document-based QA training dataset. To this end, we sourced original open-source documents, including Wikipedia, to serve as the foundation for our dataset. We employed manual annotation to ensure the generation of high-quality QA pairs from these open-source documents for our training corpus. Similarly, we constructed the evaluation dataset utilizing private documents, applying the same rigorous manual annotation process. This approach was intended to establish a dataset that could more robustly foster intricate knowledge extraction and generation capabilities within our model.

As a result of resource limitations, the public dataset we have access to is primarily composed of encyclopedias and news articles. In contrast, our private test set is mainly comprised of documents related to internet technology. There are noticeable differences between the training and test data, including proper nouns and language style. This indicates that the data falls into separate domains. Therefore, in order to obtain improved results in this cross-domain task, the model must have strong generalization capabilities.

\textbf{Train Set}. The training set consists of 143,846 documents sourced from encyclopedias and news articles, each paired with labeled question-answer pairs. 

Among these, 369 documents are deemed invalid due to their lack of sufficient information to generate meaningful questions and answers. This could be attributed to the presence of excessive URL links or unintelligible characters within those documents. Through the data, we evaluate the model's ability to filter and reject such inadequate documents.

In addition, 28,395 documents within the set are defined as summary data, indicating that the relevant information spans across multiple paragraphs. To answer the main question, one must read the entire passage, underscoring the significance of the model's capability to comprehend longer texts rather than just snippets.

\begin{figure}[ht]
    \centering
    \includegraphics[width=1.0\linewidth]{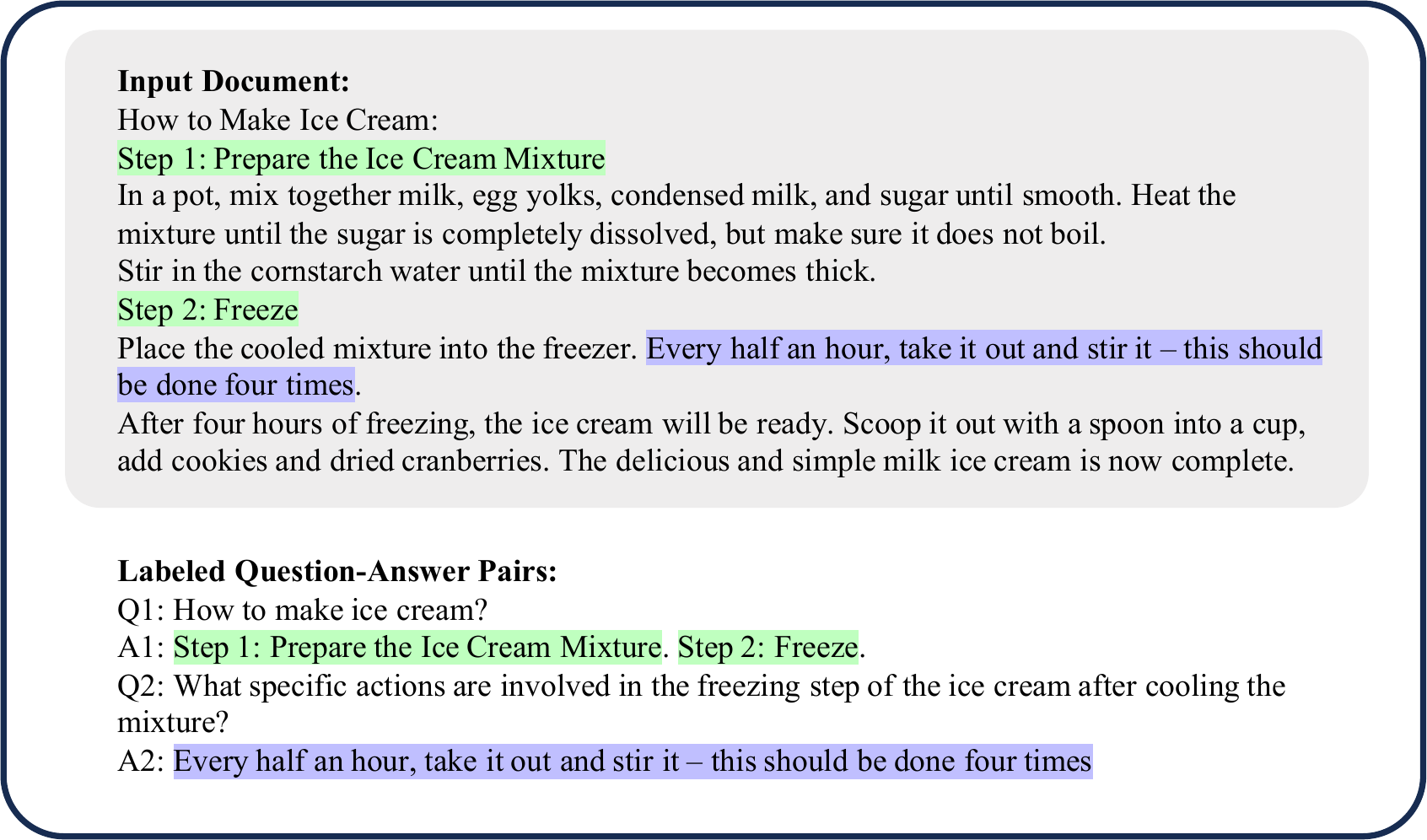}
    \caption{Example of training data}
    \label{fig:train-data}
\end{figure}

An example of the training data is shown in Figure \ref{fig:train-data}.

\textbf{Test Set}.  
The test set consists of 300 private documents that are related to Internet and technology knowledge. These documents are manually selected from challenging samples encountered during previous question-answer pair generation practice. 100 of these documents need to be rejected. The other 200 documents are summary data.

\subsection{Methods}\label{subsec3.2}
There exist numerous pre-trained LLMs that can handle all kinds of natural language processing tasks ~\citep{alpaca} including question-answer pairs generation. However, directly applying these models to specific scenarios often yields unstable output results. In order to achieve more controlled and reliable results, we have decided to fine-tune the LLMs using our carefully organized dataset mentioned in Section \ref{subsec3.1}.

\textbf{Pre-trained language models(PLMs)}

We have chosen a variety of pre-trained models, including Chinese-based models like Qwen~\citep{Bai2023QwenTR}, Baichuan2~\citep{Yang2023Baichuan2O}, ChatGLM3~\citep{zeng2022glm}, DeepSeek LLM, DeepSeek code~\citep{Bi2024DeepSeekLS}, as well as English-based models such as LLama2~\citep{Touvron2023Llama2O} and Code Llama~\citep{code_llama}. Additionally, we have developed a modified version of Code Llama.

We made modifications to the Code Llama model primarily by expanding its Chinese vocabulary list. The vocabulary size was increased from 32,016 to 40,092, encompassing an additional 8,076 commonly used Chinese words and common punctuation marks~\citep{chinese-llama-alpaca}. To accommodate the new vocabulary, we expanded the word embedding layer of the pre-trained model. We aimed to retain as much information as possible from the original word embedding layer while providing the new tokens with an initial embedding based on the average word embedding (mean embedding) of similar original tokens.

Considering our business requirements for inference efficiency and concurrency, we have primarily selected models with a size of 7 billion parameters. We conduct comprehensive experiments and evaluations of most popular Chinese models available in this size range. It is worth noting that we opt to use PLMs instead of the Supervised Fine-Tuning (SFT) versions. PLMs have achieved outstanding achievements in abstractive single-document summarization~\citep{Shen2023AHE}. And this decision was made also due to variations in the data used for SFT among these models, with some models having multiple SFT versions. These models may work well in the specific domain it was trained in, but may not generalize well to the other domain. In summary, we choose to use the PLMs to avoid any potential interference from SFT data.

\textbf{Finetuning}

We use the same data for multiple models, follow the same training data input order, and use the same hyper parameters for full or QLoRA SFT ~\citep{Dettmers2023QLoRAEF}. 

In this study, we devised four experimental protocols to evaluate the efficacy of different models in generating Chinese question-answer pairs.Experiments \ref{subsec4.2.1} entailed a comparison of the performance disparities across various metrics after SFT on code-based models versus alternative model architectures. In experiment \ref{subsec4.2.2}, we conducted a targeted comparison between code-based models in various language domains to identify differential impacts on task-specific metrics after SFT. In experiment \ref{subsec4.2.3}, we delved deeper into the impact of Chinese on the models by modifying the vocabulary. Finally, in Experiment \ref{subsec4.2.4}, we assessed the comparative effectiveness of implementing SFT versus QLoRA in the context of our task. Each experimental set was meticulously designed to isolate and analyze key factors influencing the generation of Chinese question-answer pairs, thereby providing a comprehensive overview of model capabilities within this domain.

%% file: evaluation.tex
All models mentioned above are evaluated with evaluation set described in Section \ref{subsec3.1} for both human experts evaluation and automated evaluation. We rate the knowledge extraction results of the models through multiple experts evaluate, including accuracy, knowledge completeness, refusal to answer, etc. We use the Average Expert Score(EXPERTS) to evaluate all experimental models. Our initial approach was the native Qwen 7b model utilizing few-shot method for data processing.This served as a baseline with a score of 50 out of 100. Next, we conducted SFT experiments on multiple mainstream models utilizing full parameter SFT under the same batch of training data.It is evident that all PLMs have undergone SFT, resulting in a significant improvement in the EXPERTS on this task compared to the baseline.

We have also used various automated evaluation metrics to comprehensively, quantitatively, and objectively evaluate our experimental models. Please refer to the next chapter for relevant content. 

\subsection{Evaluation Metrics Definition}

\textbf{Coverage Analysis}: We use ROUGE-L ~\citep{lin-2004-rouge} to measure the degree of content overlap and evaluate the extent to which the summary encapsulates the information presented in the source text. A higher ROUGE-L means models outputs with better coverage rate.

\textbf{Hallucination Analysis of Answers}: We define the content creation rate (CCR) , which represents the model’s tendency to hallucinate. A higher CCR score indicates more severe hallucinations.

\textbf{Completeness Analysis of Answers}: We define the average text coverage ratio (COV). COV refers to the average ratio of the longest common substring of the output text to the input text among all analysis samples. This ratio reflects the average degree of use of input text content by the output text. A higher COV score indicates a higher copy fidelity of the output text towards the input text.

\textbf{Reject Capability Analysis}: We define the model's reject capability (REJ), which indicates whether a model knows when to refuse to complete a task. A higher REJ score suggests a better ability to discern which
materials are more worthy of a knowledge generation task.

\textbf{Summarize Capability Analysis}: We define text extraction aggregation count (TEAC), representing the model’s summarizing ability, with a higher score indicating the model’s capacity to extract information from multiple places within the content as a whole. A lower TEAC score implying simpler extracted information.
We define longest increasing subsequence ratio (LISR), indicating the model’s ability to maintain consistency. A higher LISR score suggests more orderly knowledge extraction behavior.

\subsection{Results}
 
\subsubsection{Code-based LLMs better than other LLMs}\label{subsec4.2.1}

As shown in Table \ref{tab:code-based-models}, code-based LLMs generally outperform other LLMs in terms of EXPERTS. Table \ref{tab:compare-same-base} specifically highlights several of the most significant enhancements: hallucination, generalization, and rejection. The DeepSeek LLM 7b model and the DeepSeek code 7b model, which include more Chinese data in their pre-training, have the same architecture and training methods. The only difference is that the DeepSeek code 7b is mainly trained on code data~\citep{Bi2024DeepSeekLS}. Comparative analysis of various indicators shows that DeepSeek code 7b has shown improvement relative to DeepSeek LLM 7b. A comparison between Llama2 7b and Code Llama 7b, which do not contain a lot of Chinese data in their pre-training, also shows that Code Llama 7b, which includes more code data, performs better across all indicators. One possible reason why code-based LLMs perform well is that this task requires the ability to copy the original text without any faithfulness hallucination, and the code corpora can train the model's consistency ability effectively. Another reason might be that code models inherently include a lot of HTML and markdown formatted data, which are well-suited for translation tasks; our task can also be considered as a format translation task.

\begin{table}[ht]

\label{tab:code-based-models}
\begin{center}
\begin{tabular}{ccccccc}
\toprule
\multicolumn{1}{c}{\bf Pretrained Models}& \multicolumn{1}{c}{\bf EXPERTS} &  \multicolumn{1}{c}{\bf LISR}&  \multicolumn{1}{c}{\bf TEAC}&  \multicolumn{1}{c}{\bf COV}& \multicolumn{1}{c}{\bf CCR}&\multicolumn{1}{c}{\bf ROUGE-L} \\
\midrule
         Llama2 7b&  85&  56.09&  77&  51.45&  1.76& 56.29\\   
         ChatGLM3 6b&  76&  56.70&  62&  52.59&  1.53& 56.84\\   
         DeepSeek LLM 6.7b&  78&  54.18&  72&  53.38&  1.82& 56.64\\ 
         qwen 7b&  75&  55.42&  77&  51.50&  1.63& 52.72\\ 
\midrule
 DeepSeek code 6.7b& 92& 55.35& \bf 85& \bf 53.43& 1.71&55.80\\  
 Code Llama 7b& \bf 97& \bf 58.18& 77& 52.23& \bf 1.42&55.79\\  
 modified Code Llama 7b&\bf 97& 55.27& 84& 53.27& 1.49&\bf 58.51\\
\bottomrule
\end{tabular}
\end{center}
\caption{Overall, code-based LLMs tend to be preferred over other LLMs in the benchmarks studied. The difference is especially pronounced in the EXPERTS judged by humans.}
\end{table}

\begin{table}[ht]

\label{tab:compare-same-base}
\begin{center}
\begin{tabular}{ccccccc} 
\toprule
\multicolumn{1}{c}{\bf Pretrained Models}&\multicolumn{1}{c}{\bf TYPE}&\multicolumn{1}{c}{\bf EXPERTS}  &\multicolumn{1}{c}{\bf REJ}&  \multicolumn{1}{c}{\bf LISR}& \multicolumn{1}{c}{\bf TEAC} &\multicolumn{1}{c}{\bf CCR} \\
\midrule
\multirow{2}{*}{Llama2 7b}&    Raw LM&85&  82&  56.09& \bf 77& 1.76\\ 
    &    Code LM&\bf 97& \bf  89&  \bf 58.18& \bf  77& \bf 1.42\\ 
\midrule
\multirow{2}{*}{DeepSeek  6.7b}&    Raw LM&78&  84&  54.18&  72& 1.82\\ 
    &    Code LM&\bf 92&  \bf 89& \bf  55.35&\bf   85& \bf 1.71\\ 
\bottomrule
\end{tabular}    
\end{center}
\caption{Comparison of metrics between Raw LM and Code LM}
\end{table}

\subsubsection{Less Domain Knowledge, Better Performance}\label{subsec4.2.2}

Table \ref{tab:less-more-chinese} presents the domain-specific evaluation results for DeepSeek code 6.7b and Code Llama 7b, both models, which are primarily trained on code data, have nearly the same number of parameters. In accordance with the descriptions provided in their papers, the DeepSeek model ~\citep{Bi2024DeepSeekLS} is anticipated to have a considerably higher proportion of Chinese content within its training corpora compared to the CodeLlama model. Upon evaluation, it was determined that both models exhibit competency across our suite of experimental metrics, each demonstrating distinct strengths and weaknesses across various indicators. The main reason DeepSeek code 6.7b scores lower than Code Llama 7b on EXPERTS is that DeepSeek code 6.7b's responses sometimes include some information not found in the source material. This will result in some of our unique internal terms in the test set being mistaken by LLM and replaced with common vocabulary. The CCR metric also confirms that Code Llama 7b has less hallucination. In our task of knowledge generation, verbatim copying of the original content is essential; therefore, code-based LLMs with less Chinese training corpora common knowledge perform better. The experimental results indicate that an excess of Chinese knowledge can interfere with the task. In this domain-specific task, a more knowledgeable model may lead to increased instances of hallucination. Moreover, additional knowledge in the same linguistic context is more likely to be activated.

\begin{table}[ht]

\label{tab:less-more-chinese}
\begin{center}
\begin{tabular}{ccccccc}
\toprule
\multicolumn{1}{c}{\bf Pretrained Models}& \multicolumn{1}{c}{\bf EXPERTS} &  \multicolumn{1}{c}{\bf LISR}& \multicolumn{1}{c}{\bf TEAC}&\multicolumn{1}{c}{\bf COV}&\multicolumn{1}{c}{\bf CCR}&\multicolumn{1}{c}{\bf ROUGE-L}  \\ 
\midrule
 DeepSeek code 6.7b& 92& 55.35& \bf 85& \bf 53.43& 1.71&\bf 55.80\\  
 Code Llama 7b& \bf 97& \bf 58.18& 77& 52.23& \bf 1.42&55.79\\  
\bottomrule
\end{tabular}
\end{center}
\caption{Comparison of 6 domain-specific metrics between DeepSeek code 6.7b and Code Llama 7b}
\end{table}

\subsubsection{A Moderate Amount of Chinese is Better}\label{subsec4.2.3}

\begin{figure}[ht]
\begin{center}
    \includegraphics[width=1.0\linewidth]{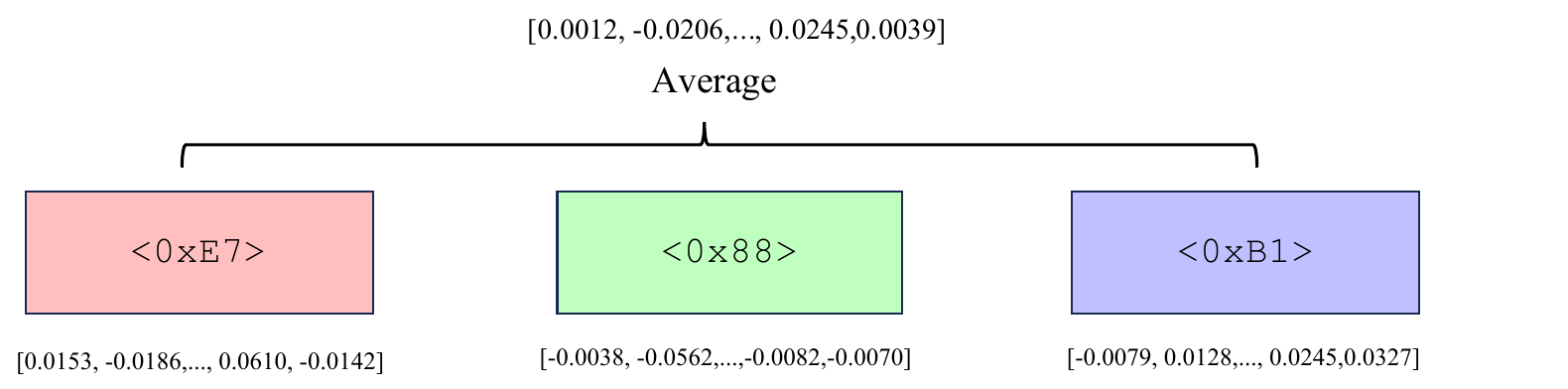}
    \caption{Embeddings of a Chinese character}
    \label{fig:love}
\end{center}
\end{figure}

Due to the conclusions of Section \ref{subsec4.2.2}, we do not want to improve the model's Chinese language abilities through additional training data. We extended LLaMA's existing vocabulary by adding additional 8076 Chinese tokens. As shown in Figure \ref{fig:love}, the embeddings for the new tokens were initialized with the average of the embeddings of the existing tokens. We expanded the vocabulary in this manner. Then we train this modified model with same data in full SFT.

On the EXPERTS metric, there was no discernible difference in the performance of the model before and after extending the LLaMA vocabulary with additional Chinese tokens and adapting the model to the extended vocabulary~\citep{chinese-llama-alpaca}. For other metrics, we observed significant improvements across the board, except for a slight degradation in the content creation rate CCR metric, as shown in Table \ref{tab:change-vocab}. 

\begin{table}[ht]

\label{tab:change-vocab}
\begin{center}
\begin{tabular}{ccccccc} 
\toprule
\multicolumn{1}{c}{\bf Pretrained Models} &\multicolumn{1}{c}{\bf REJ}&  \multicolumn{1}{c}{\bf LISR}& \multicolumn{1}{c}{\bf TEAC}&\multicolumn{1}{c}{\bf COV}& \multicolumn{1}{c}{\bf CCR}&\multicolumn{1}{c}{\bf ROUGE-L} \\ 
\midrule
 Code Llama& \bf 89&\bf 58.18& 77& 52.23&\bf 1.42&55.79\\  
 Code Llama(more embeddings)& 87& 55.27&\bf 84& \bf 53.27& 1.49&\bf 58.51\\
 Code Llama(random embeddings)& 67& 49.08& 59& 50.76& 3.28&56.76\\
\bottomrule
\end{tabular}
\end{center} 
\caption{Comparison of metrics between Code Llama 7b and Code Llama 7b with extended Chinese tokens(more embeddings) and Code Llama 7b with randomly embedding weights of Chinese tokens(random embeddings)}
\end{table}

After extending LLaMA's vocabulary by adding Chinese tokens, humans could not discern any difference in the model’s effectiveness. However, the increase in text extraction aggregation count TEAC and CCR suggest that the model's Chinese language capabilities have improved following the vocabulary extension. The enhancement in Chinese language abilities has led to improved summarizing abilities but also to more hallucinations. When the model's ability to process Chinese was completely impaired, all its metrics deteriorated significantly.

We disrupted the Chinese processing capabilities of Code Llama 7b by randomly initializing the embeddings for the newly added tokens and the existing Chinese tokens. We then found that the new model, trained in the same manner as before, performed very poorly across all metrics. From the model's output, it is evident that code-based LLMs with no understanding of Chinese are unable to learn normal Chinese conversational capabilities solely from our limited SFT data.

\subsubsection{QLoRA fails to replicate the effects}\label{subsec4.2.4}

We trained the Code Llama 7b model using two methods, scaling and noscaling, as described by~\citet{Su2021RoFormerET, Sun2022ALT}, with QLoRA, and neither approach achieved results as good as those obtained with full SFT across all metrics. As shown in Table \ref{tab:QLoRA}. Regardless of the QLoRA method, under this task, the  Code Llama 7 cannot obtain complete document QA pair generation capability through QLoRA. As described by ~\citet{Sun2023ACS}, applying LoRA-based tuning to a base model that has not been instruction-tuned shows limited effectiveness and is significantly less effective than full-parameter fine-tuning.

\begin{table}[ht]

\label{tab:QLoRA}
\begin{center}
\begin{tabular}{cccccccc}
\toprule
\multicolumn{1}{c}{\bf Pretrained Models}&\multicolumn{1}{c}{\bf EXPERTS} &\multicolumn{1}{c}{\bf REJ}&\multicolumn{1}{c}{\bf LISR}&\multicolumn{1}{c}{\bf COV}&\multicolumn{1}{c}{\bf CCR}&\multicolumn{1}{c}{\bf ROUGE-L} \\
\midrule
        Code Llama 7b FT& \bf 97& \bf 89&\bf 58.18& \bf 52.23&\bf 1.42&\bf 55.79\\  
 Code Llama 7b qlora noscaling&  28& 65& 55.43& 51.11& 1.95& 55.28\\
 Code Llama 7b qlora scaling& 26& 58& 54.10& 51.176& 1.98&54.14\\
\bottomrule
\end{tabular}
\end{center}
\caption{SFT using the QLoRA fails to replicate the effects described in this paper.}    
\end{table}

%% file: appendix.tex
\subsection{Implement details}
The hyperparameters are uniformly set to the number of train epochs=2, learning rate=1e-5, max seq length=8192, lr scheduler type=cosine, warmup steps=1000. Use a unified parameter configuration during evaluation: top p = 0.7, temperature = 0.1, repetition penalty = 1.0

\subsection{Coverage Analysis of Textual Summaries in Source Material Code}

We first refine the Longest Common Substring (LCS) function to ensure that the resulting common substrings do not contain the special character asterisk (*). If an asterisk is encountered within a candidate substring, the function iteratively seeks the next longest substring void of asterisks, continuing this process until an appropriate LCS is identified without the special character.

Subsequently, we apply the revised LCS algorithm iteratively between two strings: the summary and the original text. With each iteration, the LCS identified is replaced by asterisks in both strings to avert subsequent overlaps. This procedure is repeated until the LCS identified falls below the threshold length of four characters.

Finally, by analyzing the positions of these iteratively extracted LCSs, we determine the scope of coverage that the extracted summary has within the source material. This methodology quantitatively measures the degree of content overlap and evaluates the extent to which the summary encapsulates the information presented in the source text.

\begin{verbatim}
def longest_common_substring(str1, str2):
    # Create a 2D array to save intermediate results
    dp = [[0] * (len(str2) + 1) for _ in range(len(str1) + 1)]
    
    # Initialize the length
    # ending index of the longest common substring
    max_length = 0
    end_index = 0
    
    # Populate the dynamic programming table
    for i in range(1, len(str1) + 1):
        for j in range(1, len(str2) + 1):
            if str1[i - 1] == str2[j - 1]:
                # When encountering '*', reset dp[i][j] to 0
                if str1[i - 1] == '*':
                    dp[i][j] = 0
                else:
                    dp[i][j] = dp[i - 1][j - 1] + 1
                    # We need to check if the substring includes '*'
                    # if not, update the longest substring info
                    if (dp[i][j] > max_length 
                    and '*' not in str1[i - dp[i][j]: i]):
                        max_length = dp[i][j]
                        end_index = i
            else:
                dp[i][j] = 0
    
    # Return the longest common substring
    return str1[end_index - max_length: end_index]
    
\end{verbatim}

\begin{verbatim}
def find_largest_common_substring_bounds(source_text, one_answer):
    start_pos = None
    end_pos = None
    origin_text = source_text

    while True:
        # Find the longest common substring
        common_substring = longest_common_substring(
            source_text, one_answer)

        # Break the loop 
        # if the length of the common substring is less than N
        if len(common_substring) <= N:
            break

        # Find the new positions of the common substring 
        # in the source_text
        start_pos_new, end_pos_new = find_substring_positions(
            common_substring, origin_text)

        # Replace the common substring 
        # in both source_text and one_answer with an asterisk
        source_text = source_text.replace(common_substring, '*')
        one_answer = one_answer.replace(common_substring, '*')

        # Update the start and end bounds 
        # for the largest common substring found
        start_pos = start_pos_new if start_pos is None else min(
            start_pos_new, start_pos)
        end_pos = end_pos_new if end_pos is None else max(
            end_pos_new, end_pos)
\end{verbatim}

\subsection{Display of optimal model generation results}
Table \ref{tab:Generation results display} showcases the optimal model generation results.

\begin{table}[ht]

\label{tab:Generation results display}
\begin{center}
\begin{tabular}{ccp{9cm}}  
\toprule
\multicolumn{1}{c}{\bf Type}  &\multicolumn{1}{c}{\bf Pair} &\multicolumn{1}{c}{\bf Data} \\
\midrule
\multirow{2}{*}{To Extract}& raw input & \begin{CJK*}{UTF8}{gbsn}心智模型是对某人关于某事在现实世界中如何运作的思考过程的解释。它是周围世界、其各个部分之间的关系以及一个人对自己的行为及其后果的直觉感知的表示。心智模型可以帮助塑造行为并设置解决问题的方法（类似于个人算法）和执行任务。\textbackslash n心智模型是一种外部现实的内部符号或表示，假设在认知、推理和决策中起主要作用。肯尼斯·克雷克 (Kenneth Craik) 在 1943 年提出，大脑构建了现实的“小规模模型”，用于预测事件。\textbackslash n1971年，Jay Wright Forrester 将一般心智模型定义为：[1]\textbackslash n我们脑海中的世界形像只是一个模型。在他的头脑中，没有人想像整个世界、政府或国家。他只选择了概念和它们之间的关系，并用它们来表示真实的系统。\textbackslash n在心理学中，术语“心智模型”有时用于泛指心智表征或心智模拟。在其他时候，它被用来指心理模型和推理（英语：Mental models and reasoning）以及由 Philip Johnson-Laird 和 Ruth M.J. Byrne 开发的推理心理模型理论。\end{CJK*}\\  
         & model output & \begin{CJK*}{UTF8}{gbsn}1： 
问题：什么是心智模型？ 
答案：心智模型是对某人关于某事在现实世界中如何运作的思考过程的解释。它是周围世界、其各个部分之间的关系以及一个人对自己的行为及其后果的直觉感知的表示。 
\textbackslash n2： 
问题：心智模型的作用是什么？ 
答案：心智模型可以帮助塑造行为并设置解决问题的方法，类似于个人算法，并执行任务。 
\textbackslash n3： 
问题：谁首次提出了心智模型？ 
答案：肯尼斯·克雷克在1943年提出了心智模型，他假设大脑构建了现实的“小规模模型”，用于预测事件。 
\textbackslash n4： 
问题：Jay Wright Forrester是如何定义心智模型的？ 
答案：Jay Wright Forrester在1971年将一般心智模型定义为：我们脑海中的世界形像只是一个模型。在他的头脑中，没有人想像整个世界、政府或国家。他只选择了概念和它们之间的关系，并用它们来表示真实的系统。 
\textbackslash n5： 
问题：在心理学中，心智模型有哪些不同的含义？ 答案：在心理学中，术语“心智模型”有时用于泛指心智表征或心智模拟。在其他时候，它被用来指心理模型和推理，以及由Philip Johnson-Laird和Ruth M.J. Byrne开发的推理心理模型理论。\end{CJK*}\\
\midrule
         \multirow{2}{*}{To Reject}&raw input & \begin{CJK*}{UTF8}{gbsn}特征描述 未予评估（IUCN3.1）评估（IUCN3.1）未予评估（IUCN3.22222222未予评估（IUCN3.1）未予评估（IUCN3.1）weiyu位于平谷未予评估（IUCN3.1） 生存现状 未予评估（IUCN3.1）评估（IUCN3.1）未予评估（IUCN3.22222222未予评估（IUCN3.1）未予评估（IUCN3.1）weiyu位于平谷未予评估（IUCN3.1） 保护状况 未予评估（IUCN3.1）评估（IUCN3.1）未予评估（IUCN3.22222222未予评估（IUCN3.1）未予评估（IUCN3.1）weiyu位于平谷未予评估（IUCN3.1）\end{CJK*}\\ 
         & model output & \begin{CJK*}{UTF8}{gbsn}	对不起，这段文本没有提供足够的信息来抽取QA对。请提供更多的内容。\end{CJK*}\\
\bottomrule
\end{tabular}
\end{center}
\caption{Generation results display.}
\end{table}